\documentclass[10pt,twocolumn]{article}

\usepackage{times}
\usepackage{epsfig}
\usepackage{float}
\usepackage{graphicx}
\usepackage{amsmath}
\usepackage{amssymb}
\usepackage{dsfont}
\usepackage[section]{placeins}
\usepackage{tabularx}
\usepackage{stfloats}

\usepackage[pagebackref=true,breaklinks=true,colorlinks,bookmarks=false]{hyperref}

\begin{document}

\title{Predicting Stock Movement with BERTweet and Transformers}

\author{Michael Charles Albada\\
Georgia Institute of Technology\\
{\tt\small albada@gatech.edu}
\and
Mojolaoluwa Joshua Sonola\\
Georgia Institute of Technology\\
{\tt\small jsonola6@gatech.edu}
}

\maketitle

\begin{abstract}
   Applying deep learning and computational intelligence to finance has been a popular area of applied research, both within academia and industry, and continues to attract active attention. The inherently high volatility and non-stationary of the data pose substantial challenges to machine learning models, especially so for today's expressive and highly-parameterized deep learning models. Recent work has combined natural language processing on data from social media to augment models based purely on historic price data to improve performance has received particular attention. Previous work has achieved state-of-the-art performance on this task by combining techniques such as bidirectional GRUs, variational autoencoders, word and document embeddings, self-attention, graph attention, and adversarial training. In this paper, we demonstrated the efficacy of BERTweet, a variant of BERT pre-trained specifically on a Twitter corpus, and the transformer architecture by achieving competitive performance with the existing literature and setting a new baseline for Matthews Correlation Coefficient on the Stocknet dataset without auxiliary data sources.
\end{abstract}

\section{Introduction}

Predicting stock market movements has attracted the attention of investors and academic researchers since at least 1973 \cite{Krolak}. The efficient market hypothesis \cite{Malkiel} is commonly accepted in finance around how different types of data are disseminated and used by the market to come up with the price of financial securities. The weakest form of this hypothesis states that all past price information is priced into the current price of the security. The semi-strong form of this hypothesis claims that all public information is priced into the current price of the security. And the strongest form of this hypothesis is that all information, both public and private, is priced into the current price of the security. In order to be able to predict price movements of stocks, we must have historical price movements, public information about the stock, private information about the stock, or some combination the above. This hypothesis suggests it is more likely that the price of the financial security includes historical price movements and not public information.

 With continued advancements in this problem area, financial advisors, investors, and members of the financial community can make better investment decisions to increase their revenues and potentially allocate capital more efficiently.  This has could have a direct benefit on all members around the world with a financial stake in funds that use these techniques. In addition to its potential for profit, substantial investment from algorithmic trading firms, and potential for increasing the efficiency of the allocation of capital, predicting stock market motion is a challenging and interesting problem due to its technical properties. It combines high-dimensional time series data, complex interactions between features, challenging nonstationarity, and a high signal-to-noise ratio. Contributions to these challenging areas of machine learning have the potential to advance the field.

There are many ways that this problem is solved today by members of the financial community. Some set of peers use solely historical price data when trying to predict for future price movements. They use an array of different machine learning techniques to try to solve this problem, including neural networks, decision trees, random forests, and linear regression, among others. To further boost these methods, they use technical indicators - methods such as Bollinger Bands or RSI that transform our price series into an indicator whether we should be buying or selling the stock. Based on the efficient market hypothesis, we know that this is not the best method to predict the future movements of stocks. Instead, we must also incorporate public information into our price movements and not just historical price movements.

With the rise of deep learning and NLP methods, the financial community has begun to increasingly rely on scraping and parsing public information from public sources, especially twitter. Many models parse large subsets of tweets to infer public sentiment. These techniques allow us to take into account both historical price movements as well as public information about the stock. Many of the incremental changes in the literature improve the techniques for parsing public data and combining this information with historical price movements to extract and find price predictions.

In this paper, we build upon prior work by applying BERTweet to embed the market information from Tweets, then evaluate several model architectures, including several transformer variations, to evaluate their performance. We are predicting whether a stock will move up or down on a given day. To solve this problem, we look at prices in the previous 5 days as well as tweets in the previous 5 days.

\section{Related Work}

One of the well-studied tasks in this area is the binary prediction of whether a given stock's price will increase on the following day above a given threshold. Nguyen et al. applied a generative probabilistic topic sentiment Latent Dirichlet Allocation model on tweets for the task \cite{Nguyen2015}. As far as we are aware, Selvin et al.  were the first to apply RNN, LSTM, and sliding-CNN architectures to this particular task \cite{Selvin}. Hu et al. then introduced bidirectional gated recurrent units and multiple attention layers \cite{Hu}. 

Xu et al. added GloVe word embeddings, GRUs, and a deep generative model with temporal auxiliary targets to achieve state-of-the-art results in Stocknet \cite{Xu}. Fuli Feng et al. \cite{Feng} replaced GRUs with LSTMs and included adversarial samples in training. Chen et al. \cite{Chen} introduced auxiliary market data to this task with a graph convolutional neural network to learn relationships between corporations. Kim et al. set a new benchmark with hierarchical graph attention network to selectively propagate embedded price information among similar stocks from auxiliary data\cite{Kim}. Most recently, Multipronged Attention Network for Stock Forecasting (MAN-SF) \cite{Sawhney} established a new state-of-the-art by initializing node vectors with a bilinear projection of price and tweet embeddings, then applying a graph attention model with edges identified in an auxiliary Wikipedia dataset to learn correlations between individual equities.

Zong et al. developed a novel architecture, Multimodal Stable Fusion with Gated Cross-Attention to integrate multimodal inputs on teh InnoStock, BigData22, CIKM18, and ACL18 datasets. \cite{Zong}

Substantial advances have been made in natural language processing and sequential modeling in recent years with the release of the Transformer architecture \cite{Vaswani} and Bidirectional Encoder Representations from Transformers \cite{Devlin}. More recently, Nguyen et al. \cite{Nguyen} released BERTweet, a pre-trained language model for English Tweets that improved benchmark performance on three Tweet NLP tasks. To our knowledge, neither BERTweet nor the Transformer architecture have been previously applied to the Stocknet dataset or to this problem space, despite achieving state-of-the-art performance on multiple natural language processing tasks and time series forecasting, respectively.

\section{Problem Formulation}

We adopt the StockNet dataset for the training and evaluation of our model architectures \cite{Xu}. Following this approach, we predict the movement of a given stock \textit{s} in a pre-selected stock collection \textit{S} for a target trading day \textit{d} using the market information \textit{M} comprising relevant social media tweets and historical prices in the window $[\textit{d} - \Delta\textit{d}, \textit{d} - 1]$ where $\Delta\textit{d}$ is a fixed lag. Following the literature, we keep this fixed lag set at 5, and slide a 5-day window along the trading days to generate samples. We predict the binary indicator

\begin{equation}
y=\mathds{1}(p_d^c>p_{d-1}^c)
\end{equation}

where $p_d^c$ represents the adjusted closing price on the target date and $p_{d-1}^c$ represents the closing precise on the trading day immediately prior. Samples are labeled according to their movement percentage relative to the prior day and are assigned positive if the movement is \( \geq0.55\% \) and negative if the movement is \( <-0.5\% \). Samples where the movement is between those values are discarded and not used for training. Given the task is binary classification, binary cross-entropy was used as the objective function, defined as:
\begin{equation}
H_p(q)=-\frac{1}{N}\sum_{i=1}^{N}(log(p(y_i))+(1-y_i)\cdot log(1-p(y_i)))
\end{equation}

The Stocknet dataset consists of financial prices from Yahoo Finance (finance.yahoo.com) and tweets from Twitter (twitter.com). The financial price data can be fetched from: \href{https://github.com/yumoxu/stocknet-dataset}{here}. The tweet dataset can be fetched here: \href{https://zenodo.org/record/3237458/files/glove.twitter.27B.50d.txt.gz?download=1}{here}.

The financial price dataset has price movements from 1/1/2014 to 1/1/2016. These prices include 8 stocks from the conglomerate sector and the top 10 stocks in size from the 8 other sectors - Basic Materials, Consumer Goods, Healthcare, Services, Utilities, Financial, Industrial Goods, and Technology. This price dataset includes the date, price at the open, high price of the day, low price of the day, closing price, the adjusted closing price (to adjust for stock splits), and the volume traded on the day. The twitter dataset is a JSON encoding of all tweets in the same time horizon - 1/1/2014 to 1/1/2016. The training set consists of 19 months of data from 1/1/2014 to 8/1/2015, the validation set consists of two months from 8/1/2015 to 10/1/2015, and the test set consists of thee months from 10/1/2015 to 1/1/2016. 

\section{Approach}

After executing the \href{https://github.com/yumoxu/stocknet-code}{open-source Stocknet code} released by Xu and Cohen \cite{Xu} and evaluating the results, we replaced the bidirectional-GRU with deep bidirectional transformers based on BERTweet. We also replaced the variational movement decoder with a transformer encoder architecture applied to the historical price data. The transformer architecture for Tensorflow was modeled from \href{https://www.tensorflow.org/text/tutorials/transformer}{the official documentation}.

\begin{figure}[b]
\begin{center}
\includegraphics[width=3cm]{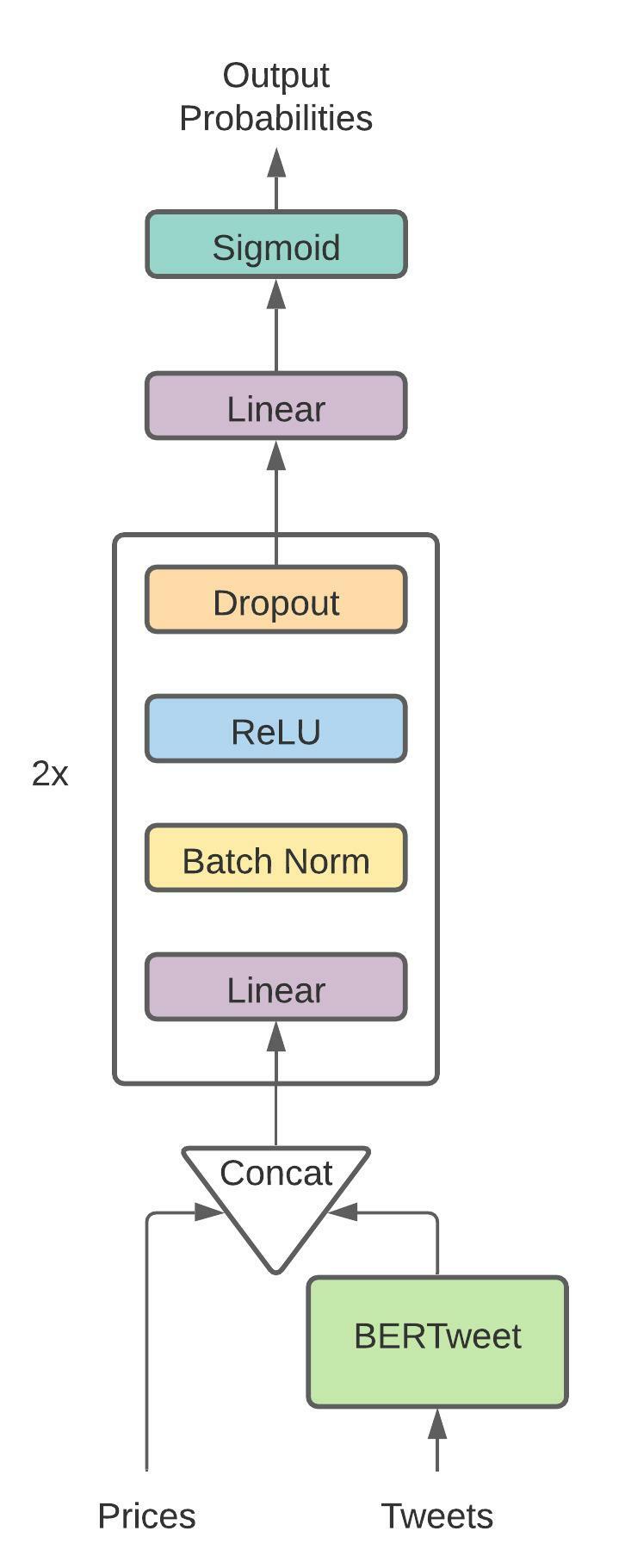}
\end{center}
   \caption{Feed Forward Network}
\label{fig:feedforwardnetwork}
\end{figure}

The original Stocknet code from 2018 is implemented in Tensorflow 1.4 and Python 2. We upgraded Stocknet to Python 3, but found upgrading Stocknet to Tensforflow 2.0 to be prohibitively time-consuming. While BERT was also supported on Tensorflow 1.11, efforts to incorporate BERT into Tensorflow 1.x's graph computation mode proved challenging, and little documentation was available, because most BERT development on Tensorflow has been on version 2.0 or above. Due to these challenges, we found switching to PyTorch allowed us to focus on model development. 

Within our PyTorch implementation, tweets were embedded using BERTweet, then concatenated with price data, and fed into one of the classifiers described below. The parameters of BERTweet were kept frozen, gradients were not backpropagated to fine-tune the BERTweet weights. This enabled pre-computation of the language model embeddings, which substantially accelerated training. The tweets were embedded to the standard dimension of 768, and all vectors for each stock for each day were averaged. The input to the model was a 3-dimensional tensor of batch size by sequence length by feature dimension. Batch size was 128, sequence length was five to match the number of days, and the feature dimension was the sum of 768 and the price feature dimension. A broad hyperparameter grid search  was run across the dimension of the feedforward layer, the key dimension for the transformers, the number of heads for multihead attention, the number of transformer layers, the dropout rate, and the auxiliary target rate. Early stopping was used to end trials early if validation loss did not decline for four epochs. Selected results are shown in the results section. Experiments were run on a cloud server with an NVidia Quadro RTX 5000 with 16 gigabytes of memory. 

The problems that we anticipated were integrating BERTweet and the transformer into the current model. While this was the main challenge, the codebase was developed on an old version of Python and Tensorflow, and bumping the codebase to more recent versions proved to be much more complex than expected and introduced thorny dependency conflicts that needed to be resolved. Finding a suitable version of Tensorflow that was compatible with BERTweet but did not require substantial changes to the Stocknet codebase was also non-trivial.

One challenge we did not anticipate was poor performance in the dataloader. The module opened and read multiple files at runtime for every sample it generated, blocking training on IO rather than on computation, which significantly slowed the pace of training and hyperparameter tuning, and required pre-fetching and indexing in appropriate data stores to support efficient sample generation. Incorporating the PyTorch data loader to enable using thread pools to parallelize data generation and pin memory for rapid access would have further accelerated training, but the nonstandard sample schema would have required substantial refactoring, and was ruled out. 

Stabilizing training also required hyperparameter tweaks. Training jobs would occasionally result in predictions of nearly all true, or nearly all false. We found that decreasing the learning rate from 1e-4 to 1e-5 and increasing the batch size from 32 to 128 led to more stable training. Adam was used as the optimizer.

\FloatBarrier
\section{Model Architecture}

Several model architectures were implemented and evaluated. The first is a simple feed-forward network trained directly on the flattened representations of the language model embedding and price data for the $[\textit{d} - \Delta\textit{d}, \textit{d} - 1]$ window. This was implemented primarily to assess the direct benefit of replacing the GloVe word embeddings combined with GRUs with a modern language model pre-trained directly on tweets. The architecture can be seen in Figure 1. 

\begin{figure}[H]
\begin{center}
\includegraphics[width=9cm]{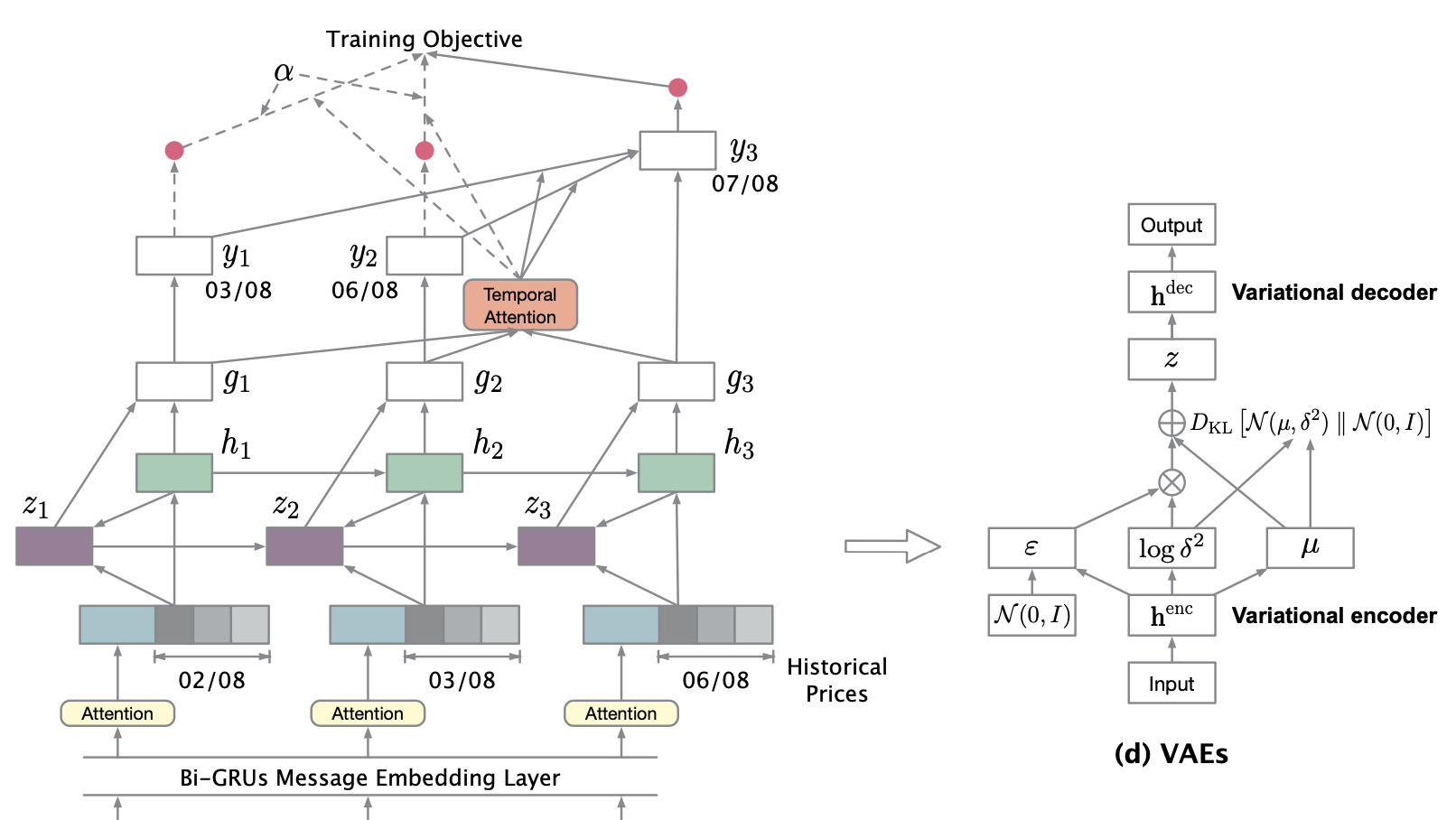}
\end{center}
   \caption{Generative Model from Xu and Cohen \cite{Xu}}
\label{fig:generativemodel}
\end{figure}

\begin{figure}[htbp]
\begin{center}
\includegraphics[width=7.5cm]{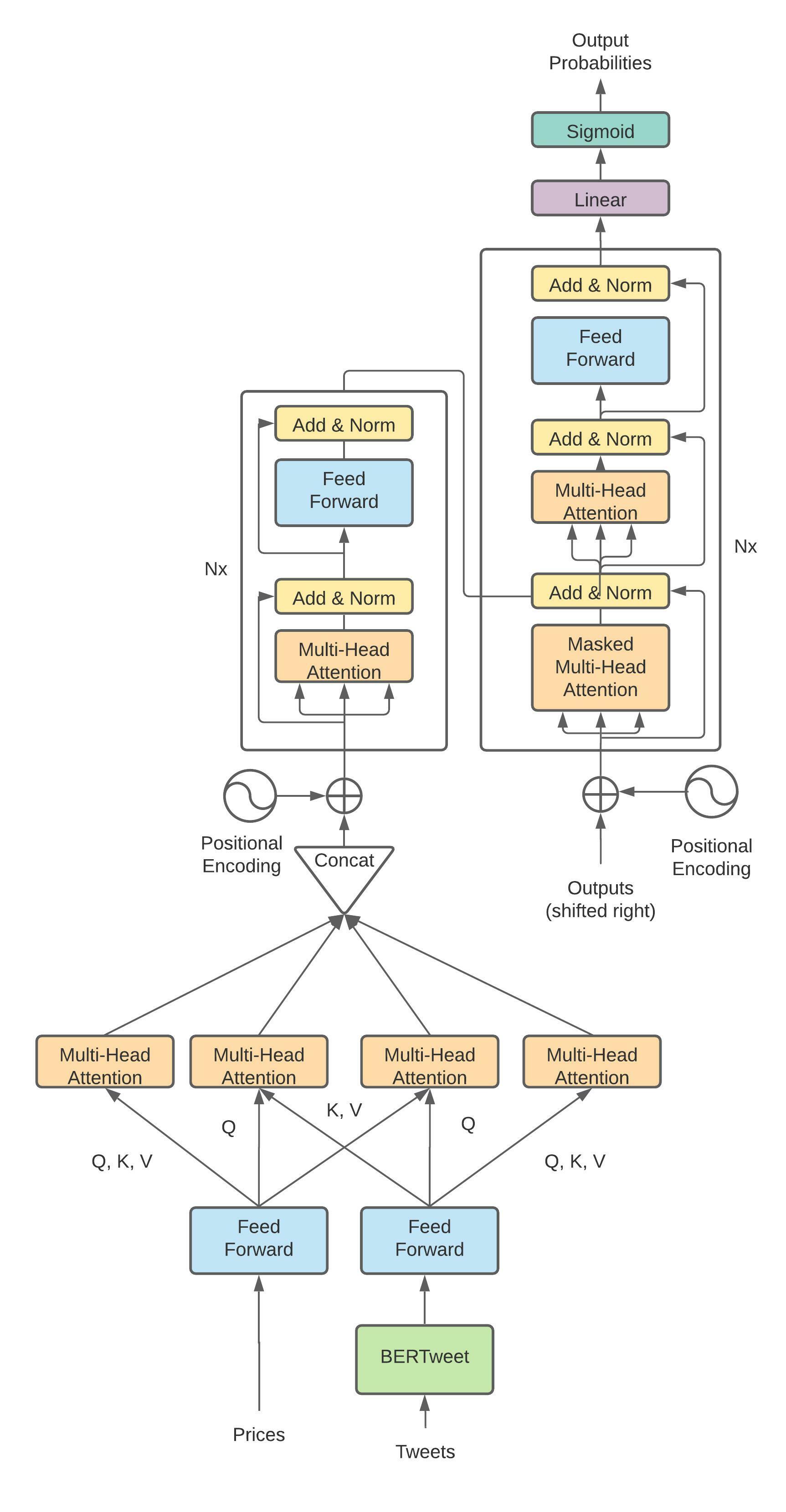}
\end{center}
   \caption{Cross-Attention Transformer with BERTweet Embedding}
\end{figure}

The second architecture replaces the variational movement decoder from the original design with a transformer. This model is similar to Figure 2 except the variational encoder/decoder is replaced with a transformer architecture for Tensorflow modeled from \href{https://www.tensorflow.org/text/tutorials/transformer}{the official documentation}. 

The third architecture is the BERTweet transformer. This design very closely follows the original transformer architecture. The primary modifications made here were applying BERTweet as the text embedding layer, concatenating this with the price data, and replacing the output softmax with a sigmoid layer to match the binary classification objective. Though the sequences were only five days in length, sinusoidal position encodings were added to represent the temporal structure of the data. This architecture more closely matches the problem because it explicitly models the time-series nature of the data while allowing the model to directly consider signal from each of the days in the window, then successive layers of multi-headed attention enable learning sophisticated interaction effects across input dimensions and time. 

The fourth architecture implemented is the cross-attention transformer. Inspired by the use of the bilinear combination in MAN-SF for combining different signals, this architecture projects both prices and tweet embeddings to a common dimension, applies self-attention independently to the price and text embeddings, then applies cross attention between the two: one in which the price projection serves as the query and the tweet project serves as the key and value, and another in which the tweet projection serves as the query, and the price projection serves as the key and value. The output from these four attention mechanisms are then concatenated, and passed through a transformer. This builds upon the prior design by explicitly representing the semantic difference between the price and tweet embeddings, which were concatenated together in the above design. The cross-attention blocks enable the model to much more directly detect complex and subtle interaction effects between the price and tweet embeddings. 

Finally, inspired by Xu and Cohen's \cite{Xu} use of temporal auxiliary targets, the transformer architectures were extended as the auxiliary transformer and auxiliary cross-attention transformer, both of which recurrently predict price movements for each of the five days. For training, gradients are computed for the auxiliary targets, and are  down-weighted gradients by the hyperparameter $\alpha$, which exists on the range [0.0, 1.0]. For inference, the auxiliary predictions are discarded, and the final prediction is used. For the target date,  $\alpha_d$ is considered to be 1, and the gradient is unadjusted. For other days in the range [d-$\Delta$ d, d], $\alpha_d$ is multiplied by the binary cross entropy for that day, before the sum is taken across each of the days in the range. More formally, the objective function becomes:

\begin{equation} 
H_p(q)=-\frac{1}{N}\sum_{d=0}^{\Delta d}\alpha_d\cdot\sum_{i=1}^{N} log(p(y_{id}))+(1-y_{id})\cdot log(1-p(y_{id}))
\end{equation}

\begin{figure}[H]
\begin{center}
\includegraphics[width=7cm]{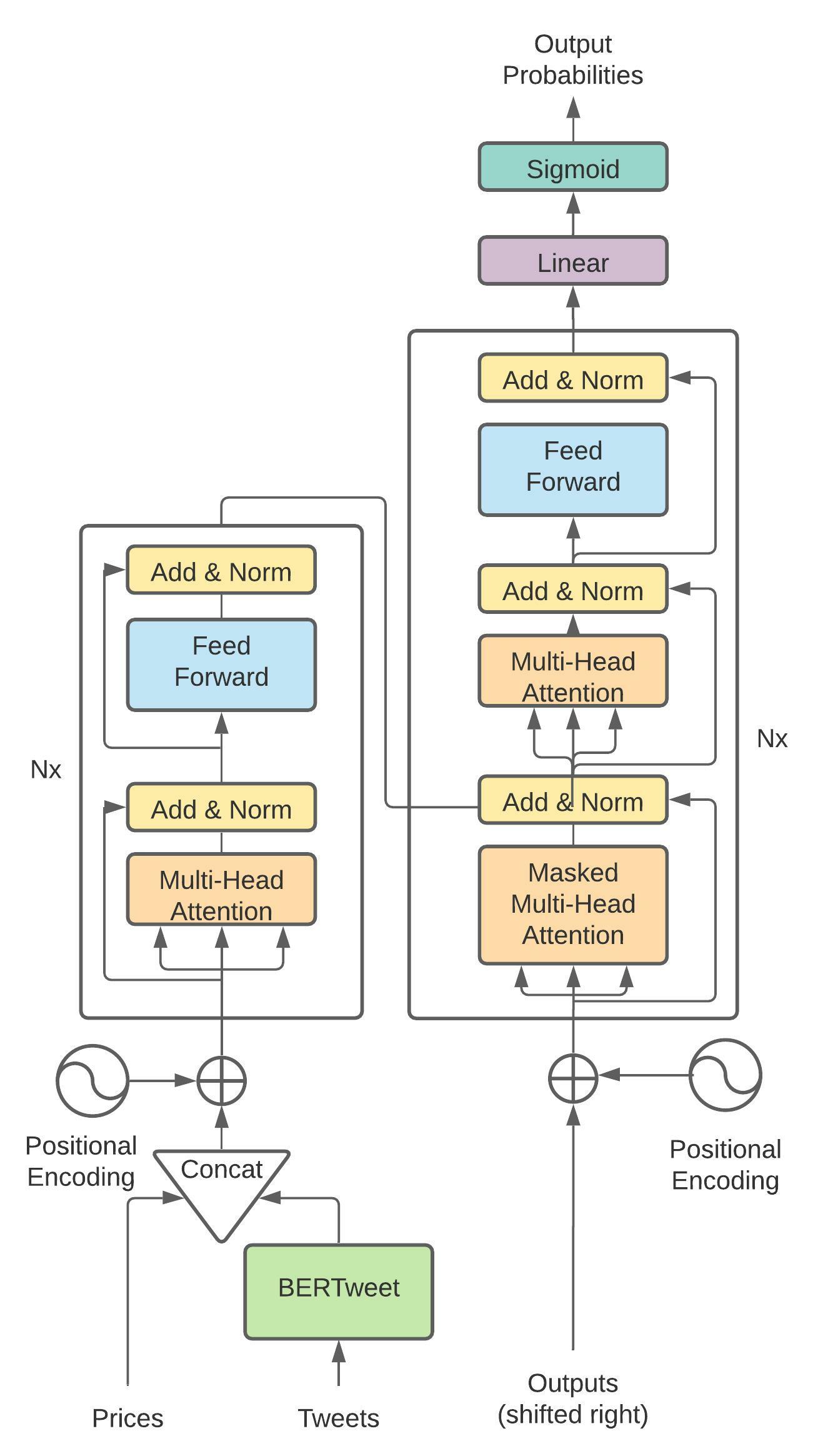}
\end{center}
   \caption{Transformer with BERTweet Embedding}
\end{figure}

\section{Experiments and Results}

\begin{table*}[b]
\begin{center}
\begin{tabular}{|l|l|l|l|l|l|l|c|l|}
\hline
Model & N & h & dim\_ff & dim\_key & dropout & aux\_weight & Accuracy & MCC \\
\hline
Generative Model Transformer & 1 & 2 & 2048 & 512 & 0.3 & & 56.25 & 0.0441 \\
Feedforward &  &  & 128 & & 0.0 & & 49.307 & 0.00788 \\
Feedforward &  &  & 1024 &  & 0.0 & & 51.11 & 0.0546 \\
Auxiliary Transformer & 8 & 2 & 2048 & 512 & 0.3 & & 51.27 & 0.06265 \\
Auxiliary Transformer & 16 & 2 & 2048 & 256 & 0.3 & & 54.09 & 0.05513 \\
Cross-Attention & 1 & 2 & 2048 & 32 & 0.0 & & 56.282 & 0.07395 \\
Cross-Attention & 1 & 2 & 2048 & 128 & 0.0 & & 52.446 & 0.0681 \\
Cross-Attention & 6 & 16 & 2048 & 512 & 0.0 & & 48.728 & 0.10070 \\
Cross-Attention & 1 & 8 & 2048 & 256 & 0.0 & & 48.336 & 0.1114 \\
Auxiliary Cross-Attention & 1 & 2 & 2048 & 512 & 0.0 & 0.5 & 56.28 & 0.07395 \\
Auxiliary Cross-Attention & 1 & 2 & 2048 & 512 & 0.0 & 0.5 & 56.634 & 0.07748 \\
Auxiliary Cross-Attention & 2 & 2 & 2048 & 512 & 0.0 & 0.5 & 56.908 & 0.08627 \\
Auxiliary Cross-Attention & 1 & 8 & 2048 & 1024 & 0.0 & 0.3 & 55.733 & 0.09264 \\
\hline
\end{tabular}
\end{center}
\caption{Accuracy and MCC by model architecture and parameters}
\label{tab:results}
\end{table*}

\begin{table*}[t]
\begin{center}
\begin{tabular}{|c|c|c|c|c}
\hline
Authors & Model & Accuracy & MCC \\
\hline
 & RAND & 50.89 & -0.002266 \\
 Nguyen \& Shirai \cite{Nguyen2015} & TSLDA & 54.07 & 0.065382 \\
 Hu et al. \cite{Hu} & HAN & 57.64 & 0.0518 \\
 Xu \& Cohen \cite{Xu}  & TechnicalAnalyst & 54.96 & 0.016456 \\
 Xu \& Cohen \cite{Xu}  & FundamentalAnalyst & 58.23 & 0.071704 \\
 Xu \& Cohen \cite{Xu}  & IndependentAnalyst & 57.54 & 0.03661 \\
 Xu \& Cohen \cite{Xu}  & DiscriminativeAnalyst & 56.15 & 0.056493 \\
 Xu \& Cohen \cite{Xu}  & HedgeFundAnalyst & 58.23 & 0.080796 \\
 Kim et al. \cite{Kim}  & HATS & 56.2 & 0.117 \\
 Chen et al. \cite{Chen} & GCN & 53.2 & 0.093 \\
 Feng et al. \cite{Feng} & Adversarial LSTM & 57.2 & 0.148 \\
 Sawhney et al. \cite{Sawhney} & MAN-SF & 60.5 & 0.195 \\
\hline
\end{tabular}
\end{center}
\caption{Accuracy and MCC from selected papers on the Stocknet dataset}
\label{tab:comparisons}
\end{table*}

Following the existing literature, we use the Stocknet evaluation framework to facilitate comparison, and only change the model architecture. Historic prices and tweets are used to predict price movements, and the model is evaluated on the test set by accuracy and Matthews Correlation Coefficient, which avoids bias due to data skew by weighting false positive, false negatives, true positives, and true negatives. All evaluations were quantitative and focused on out-of-sample prediction quality. 

\begin{equation}
MCC=\frac{tp \times tn - fp \times fn}{\sqrt{(tp + fp)(tp + fn)(tn + fp)(tn + fn)}}
\end{equation}

Replacing just the variational movement decoder with a transformer encoder architecture yielded an accuracy of 56.25 with an MCC of 0.0441. Our best results on MCC of 0.1114 were obtained using the cross-attention transformer with 8 heads and an embedding dimension of 512. Best results on accuracy of 56.908 were obtained with the cross-attention transformer with auxiliary loss with two transformers layers, two heads, an embedding dimension of 512, and an auxiliary weight of 0.5. The auxiliary targets appear to have increased accuracy, while the single-target achieves better balanced performance as measured by MCC. The feedforward networks achieved comparable performance to several publications, suggesting that BERTweet is more efficiently encoding market information from tweets, but this was not sufficient to outperform baselines alone. The various model size hyperparameters had only a modest effect on performance, and best results were somewhat surprisingly obtained with zero dropout.

These results compare favorably with the existing literature. The cross-attention transformer MCC of 0.1114 exceeded that obtained by Xu and Cohen \cite{Xu}, as well as by Chen et al. \cite{Chen}, who incorporated external data, including market data about 3024 companies in a graph convolutional network. The MCC for of 0.1114 is the highest result obtained on the Stocknet dataset without incorporating external data or synthetic samples. These results did not match the performance of the the adversarial LSTM in Feng et al. \cite{Feng}, HATS in Kim et al. \cite{Kim}, or the current state of the art graph attention network in Sawhney et al. \cite{Sawhney}.

\section{Conclusion}

In this paper, we demonstrated the effectiveness of the BERTweet language model and the transformer architecture for stock market price prediction. We tested these new architectures on the StockNet dataset and showed that it exceeded the original StockNet models, in addition to some of the more recent work, and set a new benchmark for performance on Stocknet dataset without incorporating external data. Future directions include combining the transformer architecture, adversarial sampling, auxiliary market information, and graph attention to evaluate the combined performance of these complementary improvements. 

\begingroup
\newpage
\renewcommand{\baselinestretch}{0.98}
\endgroup
\end{document}